\documentclass[journal]{IEEEtran}

\usepackage{graphicx}
\usepackage{epsfig}
\usepackage{dcolumn}
\usepackage{bm}
\usepackage{amsmath}
\usepackage{amssymb}

\usepackage{cases}

\begin{document}

\title{Memcomputing and Swarm Intelligence}

\author{Yuriy~V.~Pershin
        and Massimiliano~Di~Ventra % <-this % stops a space
\thanks{Y. V. Pershin is with the Department of Physics
and Astronomy and USC Nanocenter, University of South Carolina,
Columbia, SC, 29208 \newline e-mail: pershin@physics.sc.edu.}% <-this % stops a space
\thanks{M. Di Ventra is with the Department
of Physics, University of California, San Diego, La Jolla,
California 92093-0319 \newline e-mail: diventra@physics.ucsd.edu.}% <-this % stops a space
%\thanks{Manuscript received September XX, 2010; revised January YY,
%2011.}
}

\maketitle

%\author{Yuriy V. Pershin}
%\email{pershin@physics.sc.edu} \affiliation{Department of Physics and Astronomy and University of South Carolina Nanocenter, University of South Carolina, Columbia, South Carolina 29208, USA}
%\author{Massimiliano Di Ventra}
%\email{diventra@physics.ucsd.edu} \affiliation{Department of Physics, University of California, San Diego, California 92093-0319, USA}

\begin{abstract}
We explore the relation between memcomputing, namely computing with and in memory, and swarm intelligence
algorithms. In particular, we show that one can design memristive networks to solve
short-path optimization problems that can also be solved by ant-colony algorithms. By employing appropriate memristive
elements one can demonstrate an almost one-to-one correspondence between memcomputing and ant colony optimization approaches.
However, the memristive network has the capability of finding the solution in one {\it deterministic}
step, compared to the {\it stochastic} multi-step ant colony optimization. This result paves the way for
nanoscale hardware implementations of several swarm
intelligence algorithms that are presently explored, from scheduling problems to robotics.
\end{abstract}


\section{Introduction} \label{sec1}

Swarm intelligence is a general term that encompasses a wide range of dynamical properties of several biological and
artificial systems \cite{beni1993swarm,Kennedy1995,bonabeau1999swarm,kennedy2001swarm,Dorigo04a,beni2005swarm,dorigo2006ant,engelbrecht2006fundamentals,garnier2007biological,blum2008swarm,karaboga2009survey,kiranyaz2014}.
It is
inspired by the self-organized behavior of some biological systems, such as ant colonies or animal herds, with collective properties that are not easily identifiable from the dynamical features of single elements alone, thus leading to emergent dynamical phenomena.

This inspiration has led to the development of many
tools, such as swarm robotics \cite{beni2005swarm}, and algorithms. In fact, a prototypical example of swarm intelligence
algorithms is the ant colony
optimization algorithm proposed by Dorigo {\it et al.} in 1991 \cite{Dorigo91a,Dorigo92a}. This algorithm is useful for a variety of
computational problems, which can be reduced to finding optimal paths through graphs, whether directed or not. Specific
examples of such problems include the shortest path, traveling salesman problem, etc. \cite{Dorigo04a}.

On the other hand, it appears that certain graph optimization problems \cite{pershin13a,pershin11d} can also be solved by memcomputing \cite{diventra13a} - a novel computing paradigm
based on the ability of physical elements with memory to store
and process information on the same physical platform. Even though the memory elements
that are envisioned to realize such a paradigm are typically resistors, capacitors and inductors with memory \cite{diventra09a},
memcomputing rests on the much more general concept of an ideal machine that is an alternative to the
Turing machine paradigm: the Universal Memcomputing Machine~\cite{traversa14a}, which is a brain-inspired
architecture composed of interacting memory cells (whether passive or active) controlled by external signals.

We have shown, for example, that memcomputing using a network of memristors
(memory resistors) requires a single step to solve the shortest path problem in a maze \cite{pershin11d}, or in any 2D geometry \cite{pershin13a}. The reason for this unprecedented computing speed is related to the intrinsic massively-parallel dynamics of the entire network, where {\it all} elements act collectively and in a self-organized manner to provide the solution.

This brief introduction already shows the deep analogy between swarm intelligence behavior of biological systems and
memcomputing. For instance, in both cases, there is a time non-locality (memory) at play, which is key to the self-organized, collective behavior of the interacting elements. In fact, some work has already
touched upon this similarity, albeit in a different form then what we will show in this paper.
In Ref. \cite{gale2012comparison}, for instance, Gale {\it et al.} have compared different ant-inspired memristor-based information gathering approaches, thus revealing another aspect of this connection. It is then natural to ask how far this analogy can be pushed, what differences can be identified, and, in view of the fact that memcomputing can be realized experimentally
using nanoscale passive devices \cite{diventra13a}, whether some of the algorithms that are the flagships of swarm
intelligence can be solved in hardware with this new computing paradigm.

The main goal of this paper is precisely to compare these two seemingly different computing approaches--one brain-like (memcomputing), the other colony-like (swarm intelligence)--and understand
their analogies and differences. We use as a test bed the prototypical ant colony optimization algorithm, and indeed
show that this could be easily realized in hardware using
networks of memristive elements. Even though other memory elements, such as memcapacitors and meminductors \cite{diventra09a} could
be equally employed, albeit in a different circuit configuration, memristive elements are the most studied so far, both
theoretically and experimentally \cite{pershin11a}, and we have thus chosen them as a starting point for this comparison. More specifically, we identify a model of memristive elements such that their
networks operate in close analogy with the computation dynamics of ant colony optimization. However, memcomputing
shows a major advantage in the solution of these optimization problems: the memcomputing network may require only a single {\it deterministic} step to find the shortest path problem solution, compared to the stochastic multi-step ant colony optimization approach.

In the following sections, we provide a detailed description of both algorithms, as well as their comparison based on few illustrative examples of the shortest path optimization. In particular, Sec. \ref{sec2} describes the main features of the ant colony optimization and memcomputing approaches. Here, we also formulate a model of current-controlled memristive system with relaxation offering a close resemblance between the memristive network dynamics and the ant colony optimization. Sec. \ref{sec3} focuses on the simplest two-path problem, which is used for an in-depth comparison of both algorithms. A more complex several-path problem is considered in Sec. \ref{sec4}. Sec. \ref{sec5} considers memcomputing with more realistic threshold-type memristive systems, and Sec. \ref{sec6} concludes.

\section{Computing models} \label{sec2}

\subsection{Ant colony optimization} \label{sec21}

The ant colony optimization algorithm is based on the stochastic propagation of multiple moving agents (ants)
through a graph \cite{Dorigo04a}. Typically, the probability for the $k$-th ant to move from node $i$ to node $j$ of the
graph is calculated using
\begin{equation}
p_{ij,k}=\frac{\tau_{ij}^\alpha \eta_{ij}^\beta}{\sum\limits_m \tau_{im}^\alpha \eta_{im}^\beta},
\label{eq:ant:probab}
\end{equation}
where $\tau_{ij}$ is the amount of pheromone deposited on $(i,j)$ edge, $\eta_{ij}$ is the inverse length of
the $(i,j)$ edge, $\alpha$ and $\beta$ are numerical parameters, and the sum in the denominator is taken over allowable transitions
from the node $i$. Note that the choice of probability~(\ref{eq:ant:probab}) resembles (locally) the Kirchhoff's current law.
Additionally, at any given step, the behavior of the moving agent (artificial ant) is based on the knowledge (memory) of the amount of
pheromone deposited at graph  edges by {\it all} previous agents.
This feature closely resembles the current flow in memristive networks as we describe below.

Moreover, at each step, the amount of pheromone is updated according to
\begin{equation}
\tau_{ij}(k+1)=(1-\rho)\tau_{ij}(k)+\nu\frac{Q}{L_k},
\label{eq:ant:tau}
\end{equation}
where the parameter $\rho$ describes the pheromone evaporation, $\nu=1$ if the $(i,j)$ edge was visited by the $k$-th ant and zero otherwise,
$Q$ is a constant and $L_k$ is the ``cost'' of the $k$-th path (typically its length).

Initially, the same amount of pheromone is deposited at all edges. According to Eq. (\ref{eq:ant:probab}), the initial probabilities mainly depend on the length ``burden'' factors $\eta_{ij}$, which are the reciprocal of the distance
to be traveled by each ant.
With time, more pheromone is deposited on paths with shorter lengths $L_k$ (see Eq. (\ref{eq:ant:tau})). This attracts more ants to such paths eventually reinforcing these ``pheromone flooded'' paths.
The problem solution thus spontaneously forms through a self-reinforcement processes by selecting the path with the shortest length associated with the largest increase of the pheromone deposition.

\subsection{Memristive networks} \label{sec22}

We now describe memcomputing with memristive networks, although, as mentioned in the introduction, other memory
elements (such as memcapacitors and meminductors~\cite{diventra09a}) could be equally employed. Memristors \cite{chua71a} or more broadly memristive systems \cite{chua76a} are resistors with memory, whose states at any given time depend on the history of signals applied.  We have recently demonstrated that a memristive processor -- a network consisting of basic units (memristors plus switches) connecting grid points -- can solve the maze and 2D shortest optimization path problems quite easily~\cite{pershin13a,pershin11d}. The ability of a network to serve as a computing machine is determined by various factors. The most important of these have been recently summarized \cite{diventra13a}, where the type and functionality of devices with memory play a key role in the network dynamics.

Taking into account the wide variety of memristive materials and devices presently known \cite{pershin11a, diventra11a} which offer a wide range of functionalities (e.g., short- and long-term memories, unipolar and bipolar behavior, etc.), we will focus in this work on rather compact device models  than on their specific material realizations.
However, these models, especially those of memristive elements with current or voltage thresholds, can be engineered and fabricated with similar features as those we use in this paper~\cite{pershin11a, diventra11a}. Therefore, while threshold-less memristive models \cite{chua71a} may offer a closer similarity to the ant colony algorithms, the more realistic memristive devices with thresholds allow us to understand practical aspects of the actual hardware realization of ant colony optimization approaches.

Memristive devices and systems can be defined either in current- or voltage-controlled form \cite{chua76a,pershin11a}. Since the ant propagation defined by Eq. (\ref{eq:ant:probab}) could be associated with a current flow, the current-controlled definition seems the most natural for our purposes, keeping in mind that current-controlled systems can, most of the times, be easily converted into voltage-controlled ones~\cite{pershin11a}. An $n$th-order current-controlled memristive system is described by
the equations
\begin{eqnarray}
V_M(t)&=&R\left(x,I,t \right)I(t) \label{eq1}\\
\dot{x}&=&f\left(x,I,t\right) \label{eq2}
\end{eqnarray}
where $V_M(t)$ and $I(t)$ denote the voltage and current across the
device, $R$ is the memristance, $x$ is a vector representing $n$ internal {\it state variables},
and $f$ is a continuous n-dimensional vector function \cite{chua76a,diventra09a}.

In order to obtain a memristive network representation of Eqs. (\ref{eq:ant:probab}), (\ref{eq:ant:tau}), the memristive network should offer a relaxation property (mimicking the pheromone evaporation). Such functionality could be
easily realized both within the internal state variables or the external sources. For the sake of simplicity, we consider here a model of memristive devices with relaxation of the internal state variables, although the second possibility -- the use of e.g., appropriate external current/voltage pulses to achieve a relaxation in the network of non-volatile memristive devices -- is also possible. Additionally, for the analogy we want
to highlight, it is of practical help to work with the inverse of the resistance -- the conductance -- since for resistors connected in parallel the current splitting is defined by conductances.

By taking into account all these preliminaries, we assume that our memristive devices are defined by
\begin{equation}
R^{-1}\left(x \right)\equiv \sigma\left(x \right)=\sigma_{on}x+\sigma_{off}\left( 1 -x\right).\label{eq3}
\end{equation}
Here, $\sigma_{on}$ and $\sigma_{off}$ are two limiting values of conductance, with $\sigma_{on}>\sigma_{off}$, and $0\leq x\leq 1$. The equation for the internal state variable includes a drift term (proportional to the current flowing through the device) and a relaxation term:
\begin{equation}
\frac{\textnormal{d} x}{\textnormal{d} t}=\kappa I(t)-\Gamma x \label{eq4},
\end{equation}
where $\kappa$ is a constant and $\Gamma$ is the relaxation rate.

In the model of the current-controlled memristive system with threshold we will employ later, the
internal state variable $x$ is described by
\begin{numcases} {\frac{\textnormal{d}x}{\textnormal{d}t}=}
-\Gamma x & for $|I|<I_t$  \label{Icontr5}
\\
\textnormal{sgn}\left( I\right)\kappa\left(\left|I\right|-I_t\right)-\Gamma x & for $|I|\geq I_t$ \;\;\;\; \label{Icontr6}
\end{numcases}
where $I_t$ is the threshold current.

Let us then consider a network of memristive systems connected to a constant current source.
Once the current bias is switched on, electrons will flow through the system, while satisfying Kirchhoff's laws. It is the presence of these laws, which are
intrinsically {\it collective}, that forces the many-body electron system to flow through the shortest path(s). In turn, these shortest paths will support the largest current, thus changing the
most, due to the presence of memory, the associated device states along them.
In other words, similar to the ant colony algorithm, it is the {\it cumulative} effect of {\it all} electrons in the
system, that makes the shortest path emerge after the initial perturbation has been switched on. This path is then
the one that self-reinforces during dynamics due to the time non-locality (memory) of the single devices. It is,
however, important to realize that in the present case, the solution may emerge in only one step of the dynamics when
the system is implemented in hardware. In addition, the memcomputing approach, as we have formulated it here, is
fully deterministic, in the sense that no transition probability of the type~(\ref{eq:ant:probab}) needs to be
imposed on the system.

\subsection{Mapping between ant colony and memristive networks} \label{sec23}

The analogy and relation between the ant colony and the memristive network to solve optimization problems is then
complete. This analogy is summarized in Table \ref{tabl1}, where for each "Nature-inspired" feature we provide
both the parameter used in the ant colony algorithm and the memcomputing approach realized with memristive elements.

However, what is not easily evident from this table is the stochastic vs. deterministic nature of the two different
approaches as discussed above. Indeed, by assuming $I_0=1$ mA current pulse of $t_0=10$ ns duration, the number of transferred electrons
is macroscopically large, namely, $N=I_0t_0/e\approx 6.2 \cdot 10^6$. Therefore, in a network of deterministic memristive devices,
the device state change will develop with a smaller contribution from each electron contrary to the typical scenario of ant optimization approach in which the role of individual moving agents is typically more important.

After this general discussion, we can now provide explicit examples of specific optimization problems.

\begin{table*}[t]
\begin{center}
\begin{tabular}{| c | c | c | }
  \hline
  Nature & Ant colony optimization & Memcomputing \\ \hline
  Natural habitat & Graph & Memristive network \\
  Ants & Artificial ants (agents)  & Electrons \\
  Length burden & The reciprocal of distance, $\eta$ & Off-state conductances \\
  Pheromones & Artificial pheromones, $\tau_i$ & Device Memory \\
  Pheromone evaporation & Artificial pheromone evaporation & Relaxation of device state \\
  Foraging behavior & Edge selection rules & Kirchoff's laws \\
     \hline
\end{tabular}
\vspace{0.2cm}
\caption{Correspondence between the ant colony optimization and memcomputing approaches.} \label{tabl1}
\end{center}
\end{table*}
% some parts are taken from http://www.math.ucla.edu/~wittman/10c.1.11s/Lectures/Raids/ACO.pdf

\section{Two-path problem} \label{sec3}
\begin{figure}[tb]
\begin{center}
\includegraphics[angle=0,width=7cm]{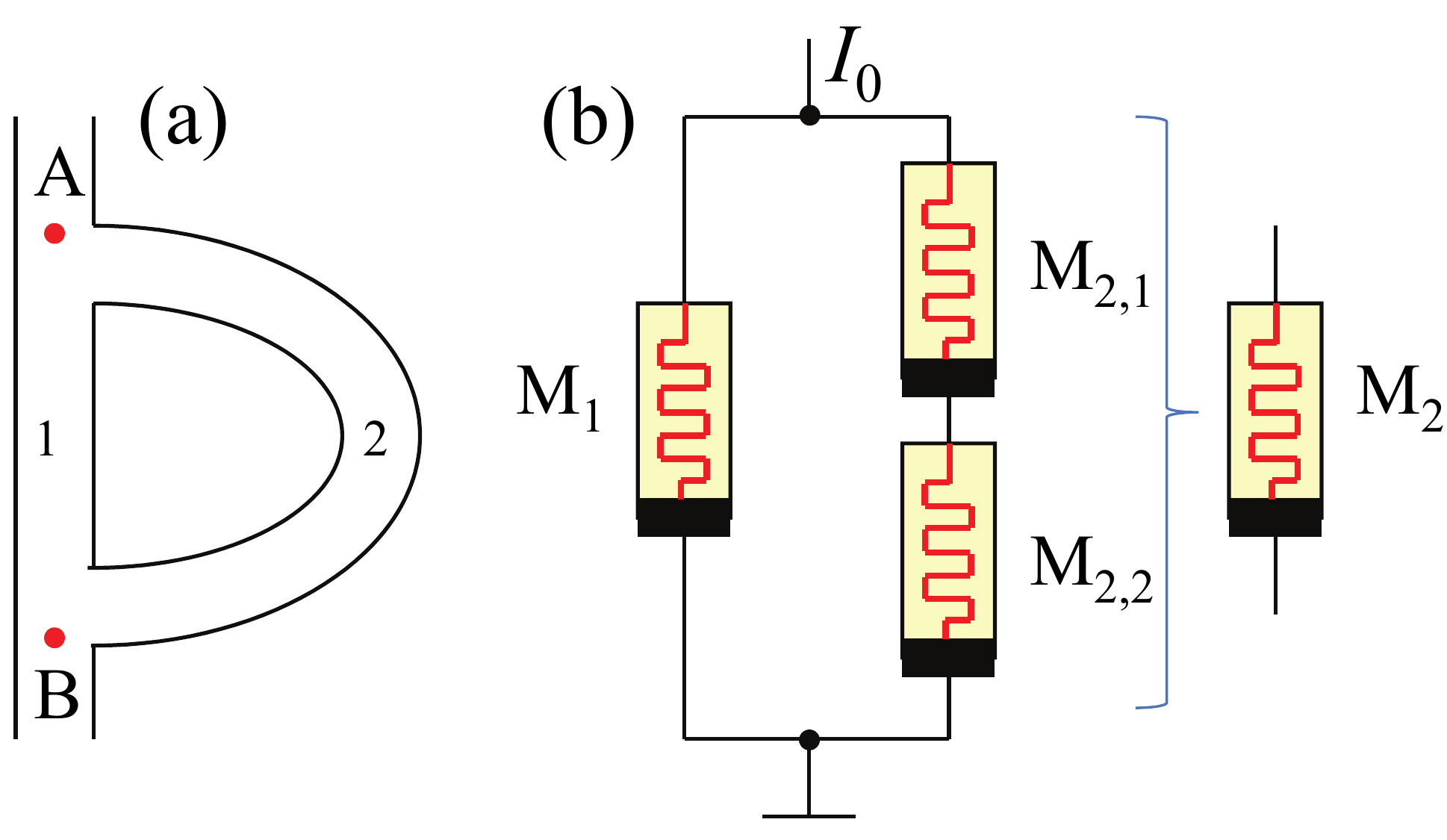}
\caption{(Color online) (a) Points A and B are connected by two paths of length $L_1$ and $L_2$, such that (for the sake of simplicity) $L_2=2L_1$. (b) Memristive network modeling the shortest path problem shown in (a).}
\label{fig1}
\end{center}
\end{figure}

\subsection{Ant colony solution}

As a first example we consider the simplest two-path optimization problem of finding the shortest path connecting two points A and B as shown in Fig. \ref{fig1}(a). It is assumed that A and B are connected by two paths of different lengths, $L_1$ and $L_2$, and, for the sake of simplicity, we choose $L_2=2L_1$. Using the ant colony optimization algorithm described in Sec. \ref{sec21}, the ants added to the point A select path $1(2)$ with probability $p_{1(2)}$ given by
\begin{equation}
p_{1(2)}=\frac{\tau_{1(2)}^\alpha \left(1/L_{1(2)} \right)^\beta}{\tau_{1}^\alpha \left(1/L_1 \right)^\beta+\tau_{2}^\alpha \left(1/L_2 \right)^\beta}
\end{equation}
The equation describing the pheromone dynamics on the first (second) path is
\begin{equation}
\tau_{1(2)}(k+1)=(1-\rho)\tau_{1(2)}(k)+\nu\frac{Q}{L_{1(2)}}.
\label{eq6}
\end{equation}

In order to make a straightforward comparison with the memristive network dynamics, we reformulate Eqs. (\ref{eq6}) in the continuous form. Let us assume that ants are added to the point A at a constant rate $\gamma$. Then, the amount of ants added within a time interval $\textnormal{d}t$ is $\gamma \textnormal{d}t$. Considering the change of $\tau_1$ and $\tau_2$ within $\textnormal{d}t$ one can find
\begin{eqnarray}
\frac{\textnormal{d}\tau_1}{\textnormal{d}t}=-\gamma\rho\tau_1&+&p_1\frac{\gamma Q}{L_1}=  \label{eq7} \\
&-&\gamma\rho\tau_1+\frac{\gamma Q}{L_1}\frac{\tau_{1}^\alpha \frac{1}{L_1^\beta}}{\tau_{1}^\alpha \frac{1}{L_1^\beta}+\tau_{2}^\alpha \frac{1}{L_2^\beta}}, \nonumber
\end{eqnarray}
and
\begin{eqnarray}
\frac{\textnormal{d}\tau_2}{\textnormal{d}t}=-\gamma\rho\tau_2&+&p_2\frac{\gamma Q}{L_2}=  \label{eq8} \\
&-&\gamma\rho\tau_2+\frac{\gamma Q}{L_2}\frac{\tau_{2}^\alpha \frac{1}{L_2^\beta}}{\tau_{1}^\alpha \frac{1}{L_1^\beta}+\tau_{2}^\alpha \frac{1}{L_2^\beta}}. \nonumber
\end{eqnarray}

Fig. \ref{fig2}(a) demonstrates the pheromone dynamics obtained as a numerical solution of Eqs. (\ref{eq7}), (\ref{eq8}).

\subsection{Memristive network solution}

In order to solve the shortest path problem shown in Fig. \ref{fig1}(a) we design a memristive network as that presented in Fig. \ref{fig1}(b). In this network, the shorter path 1 is modeled by the memristive system M$_1$ and the longer path 2 by two memristive systems M$_{2,1}$ and M$_{2,2}$ \footnote{One can use a single physical memristive device to represent path 2 encoding path 2 length in its parameters.}. Although not necessary, we take all three memristive systems to be identical and initialized in the $\sigma_{off}$ state. The memristive network is then driven by the constant current $I_0$, which splits between two possible paths according to their conductances, namely, $I_{1(2)}=I_0\sigma_{1(2)}/(\sigma_1+\sigma_2)$.

\begin{figure}[t]
\begin{center}
\includegraphics[angle=0,width=8cm]{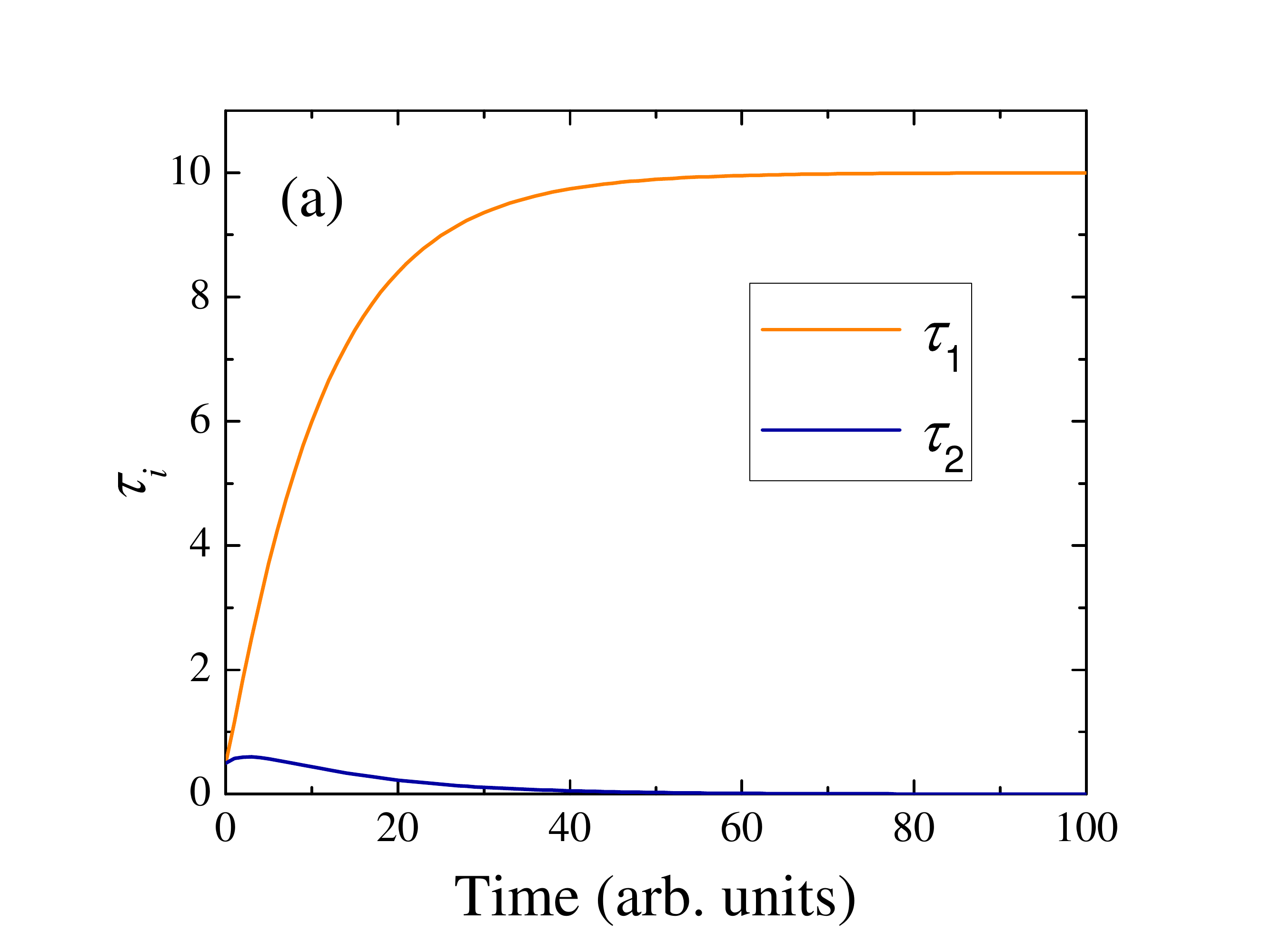}
\includegraphics[angle=0,width=8cm]{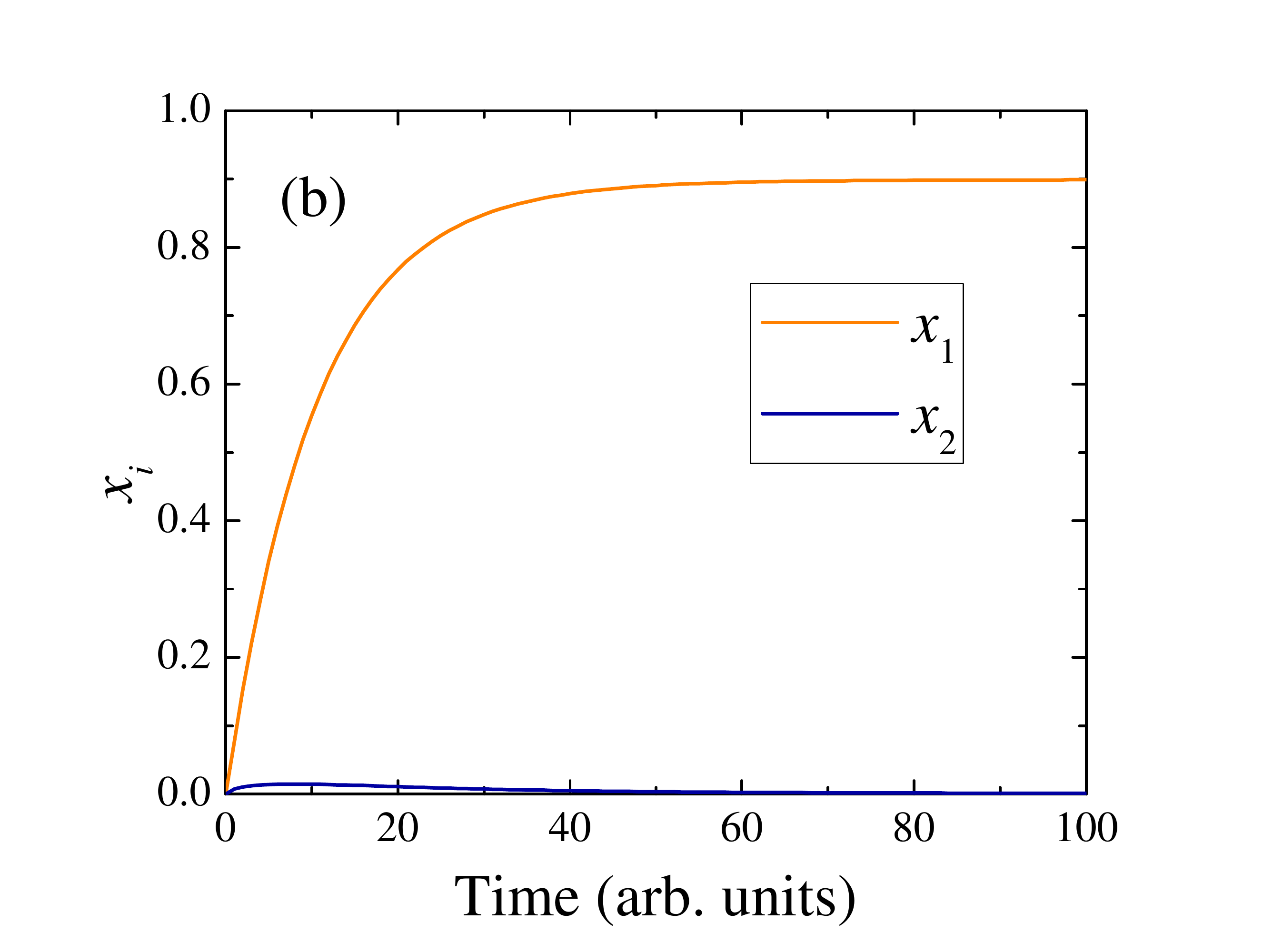}
\caption{(Color online) Time dependence of the amounts of (a) pheromone in the ant colony optimization algorithm and (b) internal state variables in the memristive network. These plots were obtained using the following parameter values: (a) $\tau_1(0)=\tau_2(0)=0.5$, $\alpha=\beta=1$, $L_1=1$, $L_2=2$, $\rho=0.1$, $\gamma=1$, $Q=1$, and (b) $\sigma_{on}=0.01$ S, $\sigma_{off}=0.00001$ S, $\Gamma=0.1$ s$^{-1}$, $\kappa=1$ (s$\cdot$A)$^{-1}$, $I_0=0.09$ A. The plot in (a) was obtained using numerical integration of Eqs. (\ref{eq7}), (\ref{eq8}).}
\label{fig2}
\end{center}
\end{figure}

Next, to render the equations more compact, we describe M$_{2,1}$ and M$_{2,2}$ as a single memristive system M$_{2}$ such that (taking into account the same parameters of M$_{2,1}$ and M$_{2,2}$ and the same initial states) its internal state variable $x_{2}=(x_{2,1}+x_{2,2})/2=x_{2,1}=x_{2,2}$ is described by Eq. (\ref{eq4}) with $\sigma_{on,2}=\sigma_{on}/2$, $\sigma_{off,2}=\sigma_{off}/2$. Using $\sigma_{i}(0)=\sigma_{off, i}$ and $\tilde \sigma_{i}=\sigma_{i}/\sigma_{off,i}$, Eq. \ref{eq4} for M$_1$ and M$_{2}$ can then be written as
%\begin{eqnarray}
%\frac{\textnormal{d} x_{1}}{\textnormal{d} t}&=&-\Gamma x_{1}+\alpha I_0 \frac{\frac{\sigma_{1}(t)}{\sigma_{1}(0)}\sigma_{1}(0)}{\frac{\sigma_{1}(t)}{\sigma_{1}(0)}\sigma_{1}(0)
%+\frac{\sigma_{23}(t)}{\sigma_{23}(0)}\sigma_{23}(0)} \label{eq9}, \\
%\frac{\textnormal{d} x_{23}}{\textnormal{d} t}&=&-\Gamma x_{23}+\alpha I_0 \frac{\frac{\sigma_{23}(t)}{\sigma_{23}(0)}\sigma_{23}(0)}{\frac{\sigma_{1}(t)}{\sigma_{1}(0)}\sigma_{1}(0)
%+\frac{\sigma_{23}(t)}{\sigma_{23}(0)}\sigma_{23}(0)}  \label{eq10} ,
%\end{eqnarray}
\begin{eqnarray}
\frac{\textnormal{d} \tilde\sigma_{1}(t)}{\textnormal{d} t}&=&-\Gamma \left[ \tilde \sigma_1-1 \right]+ \nonumber
\\
& \kappa I_0 & \left[\frac{\sigma_{on,1}}{\sigma_{off,1}} -1\right]\frac{\tilde\sigma_{1}(t)\sigma_{off,1}}{\tilde\sigma_{1}(t)\sigma_{off,1}
+\tilde\sigma_{2}(t)\sigma_{off,2}},\;\; \;\;\; \label{eq11} \\
\frac{\textnormal{d} \tilde\sigma_{2}(t)}{\textnormal{d} t}&=&-\Gamma \left[ \tilde \sigma_{2}-1 \right]+ \nonumber
\\
& \kappa I_0 & \left[\frac{\sigma_{on,2}}{\sigma_{off,2}} -1\right]\frac{\tilde\sigma_{2}(t)\sigma_{off,2}}{\tilde\sigma_{1}(t)\sigma_{off,1}
+\tilde\sigma_{2}(t)\sigma_{off,2}}.\;\; \;\;\; \label{eq12}
\end{eqnarray}

Comparing Eqs. (\ref{eq12}), (\ref{eq11}) with Eqs. (\ref{eq7}), (\ref{eq8}) one can easily notice that the memristive network we have chosen realizes the ant colony optimization algorithm with parameters $\alpha=1$, $\beta=1$ \footnote{Different values of $\alpha$ and $\beta$ could be realized with different types (models) of memristive devices}. In this realization, the initial (off-state) resistance ($1/\sigma_{off,1(2)}$) plays the role of the path length $L_{1(2)}$, and the normalized conductance $\sigma_{1(2)}(t)/\sigma_{off,1(2)}$ (proportional to the internal state variables) - the role of the pheromone strength $\tau_{1(2)}$.

However, there are also some differences. For instance, comparing relaxation terms in Eqs. (\ref{eq7}), (\ref{eq8}) and  (\ref{eq11}), (\ref{eq12}), one can notice that while $\tau_1$ and $\tau_2$ relax to zero, the memristive devices relax to a smaller but finite $\sigma_{off,i}$. Additionally,
the prefactors preceding the fractions in  Eqs. (\ref{eq11}), (\ref{eq12}) do not explicitly contain the inverse length (compared to the prefactors in Eqs. (\ref{eq7}), (\ref{eq8})). Despite these differences, simulations
clearly demonstrate a strong similarity in the time dependence of the pheromone (ant colony) and the internal state variables (memristive network), see Fig. \ref{fig2}. In both cases, the correct solution is found.

\begin{figure}[tb]
\begin{center}
\includegraphics[angle=0,width=7cm]{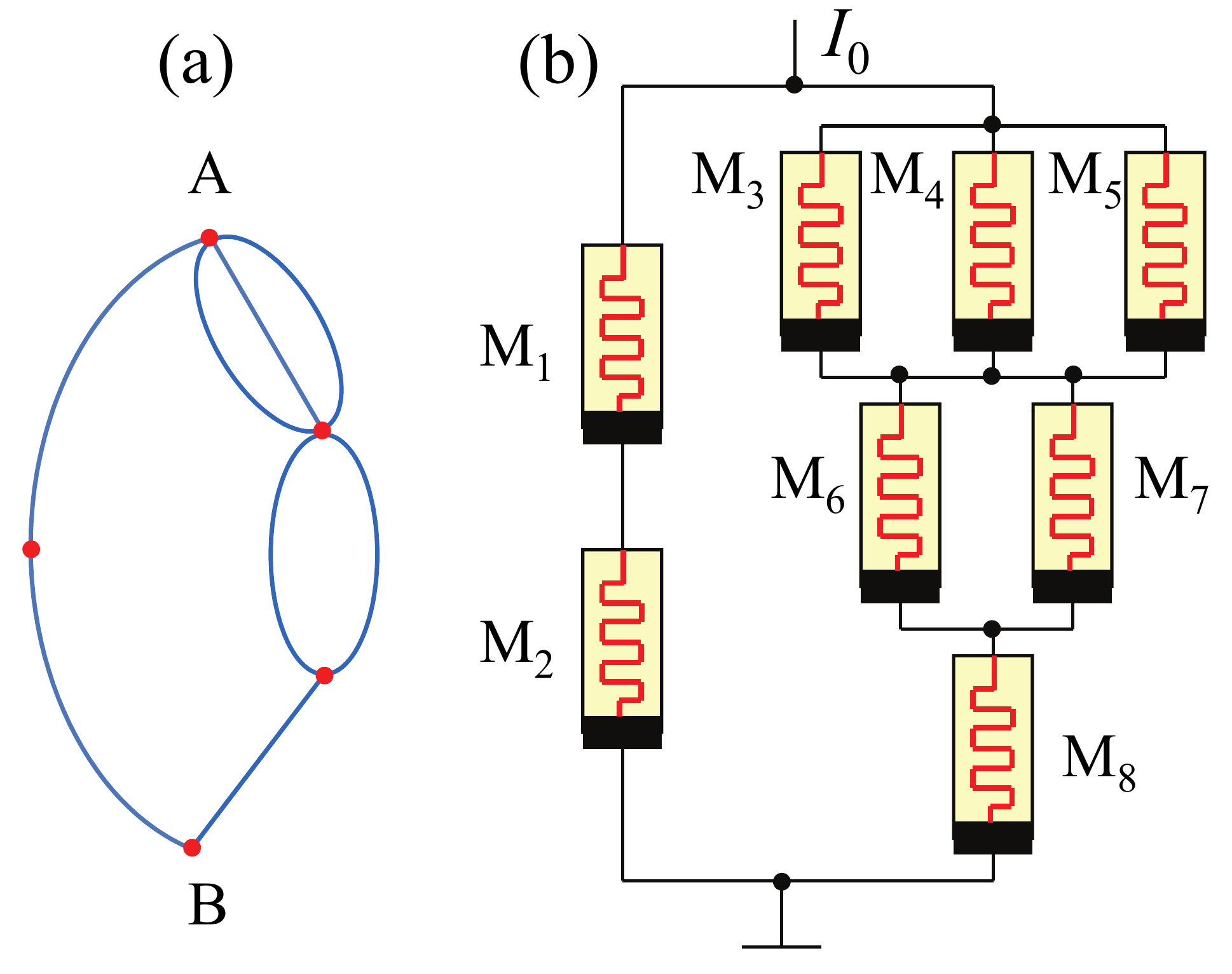}
\caption{(Color online) (a) A graph with equal weight edges. The shortest path problem is solved with respect to nodes denoted by A, B. (b) Memristive network modeling the shortest path problem shown in (a).}
\label{fig3}
\end{center}
\end{figure}

In order to get additional insight into the steady-state properties, one can easily find steady-state solutions of Eqs. (\ref{eq7}), (\ref{eq8}) and Eqs. (\ref{eq11}), (\ref{eq12}). While both pairs of equations have two solutions, the relevant solution of  Eqs. (\ref{eq7}), (\ref{eq8}) for $\alpha=\beta=1$ reads
\begin{eqnarray}
\tau_1=\frac{Q}{L_1\rho}, \label{eq13} \\
\tau_2=0   .\label{eq14}
\end{eqnarray}
Instead, the memcomputing steady-state solution of (\ref{eq11}), (\ref{eq12}) can be written as
\begin{eqnarray}
\tilde\sigma_1=\frac{C-\Gamma+\sqrt{C^2+2C\Gamma+9\Gamma^2}}{2\Gamma}, \label{eq15} \\
\tilde\sigma_2=\frac{C+5\Gamma-\sqrt{C^2+2C\Gamma+9\Gamma^2}}{2\Gamma}   ,\label{eq16}
\end{eqnarray}
where $C=\kappa I_0 \left[\sigma_{on,1}/\sigma_{off,1} -1\right]$. The specific limiting values of conductances listed above Eq. (\ref{eq11}) have been used in the derivation of Eqs. (\ref{eq15}), (\ref{eq16}). Considering $\Gamma \ll C$ limit, one gets $\tilde\sigma_1\approx C/\Gamma$ and $\tilde\sigma_2\approx 2$, which corresponds to the
values of internal state variables, $x_1\approx 0.9$ and $x_2\approx 0$. Since $C/\Gamma \gg 1$ in our
choice of parameters, this result is indeed in close agreement with the numerical solution obtained from Eqs. (\ref{eq13}), (\ref{eq14}) (see Fig. \ref{fig2}).

We emphasize again that the memcomputing solution is fully deterministic, and that it is stored into the states
of memristive systems and hence can be read directly from them. Only a single current pulse is required to solve the
problem. Although the obvious choice of the pulse duration is dictated by the time scale of reaching the steady state ($t\gtrsim 50$ in \ref{fig2}(b)),
one can notice from Fig. \ref{fig2}(b) that $x_1>x_2$ almost immediately after the current bias is switched on. Therefore,
the full time evolution is not necessarily needed to read the correct solution.
Moreover, in the actual hardware implementation, the intrinsic mechanisms of the internal state relaxation
impose limits on the smallest time scale that should be detected by the reading device.

\section{Multiple-path problems} \label{sec4}

Next, we solve the shortest path problem in a more complex graph as that presented in Fig. \ref{fig3}(a).
By inspecting this graph visually, one can easily notice that the shortest path solution consists of two edges making the left arm connecting A and B. However, considering the equivalent memristive network shown in Fig. \ref{fig3}(b), one can realize that initially, when all memristances are the same, the current is stronger in the right arm corresponding to a longer path. Despite this, the correct solution can be found using the memristive network.

\begin{figure}[tb]
\begin{center}
\includegraphics[angle=0,width=7cm]{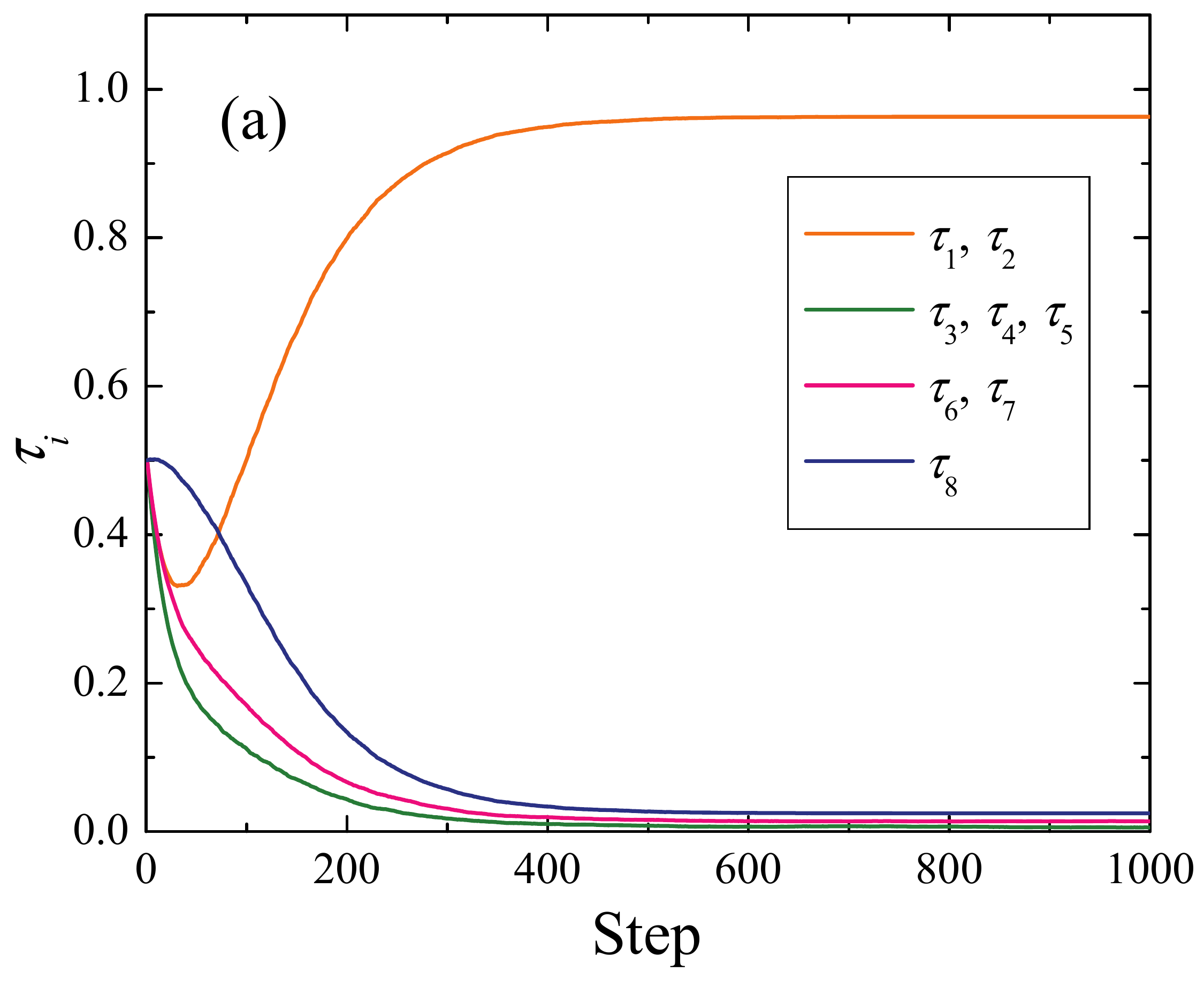}
\includegraphics[angle=0,width=7cm]{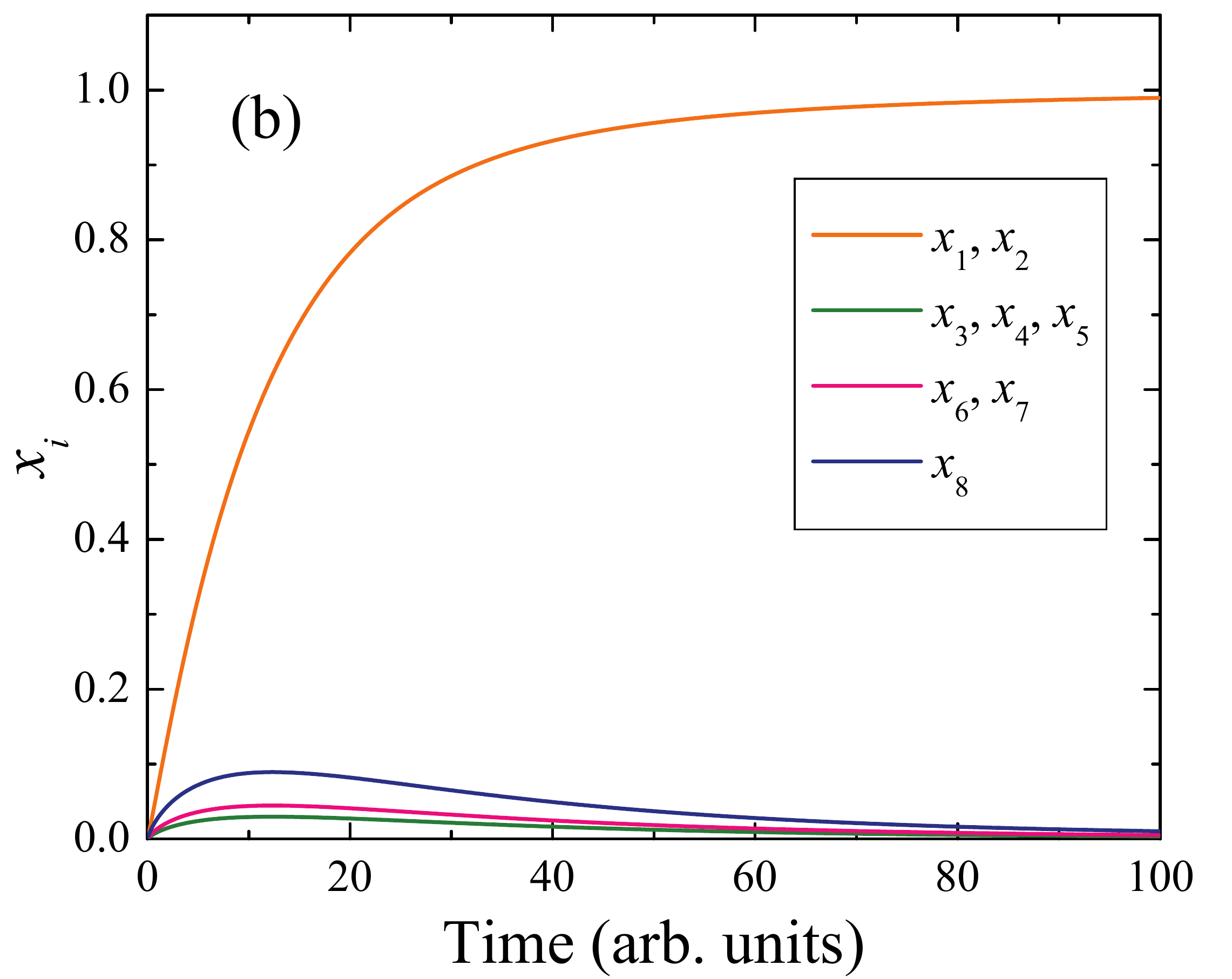}
\caption{(Color online) (a) Solution of the shortest path problem from Fig. \ref{fig3}(a) with ant colony optimization algorithm. (b) Dynamics of internal state variables in the network from Fig. \ref{fig3}(b). These plots were obtained using the following parameter values: (a) $\tau_i(0)=0.5$, $\alpha=\beta=1$, $\rho=0.05$, $Q=0.1$, and (b) $\sigma_{on}=0.01$ S, $\sigma_{off}=0.00001$ S, $\Gamma=0.1$ s$^{-1}$, $\kappa=1$ (s$\cdot$A)$^{-1}$, $I_0=0.1$ A. The plot in (a) was obtained by averaging over $10^3$ realizations of $10^3$ ants.}
\label{fig4}
\end{center}
\end{figure}

\begin{figure}[tb]
\begin{center}
\includegraphics[angle=0,width=7cm]{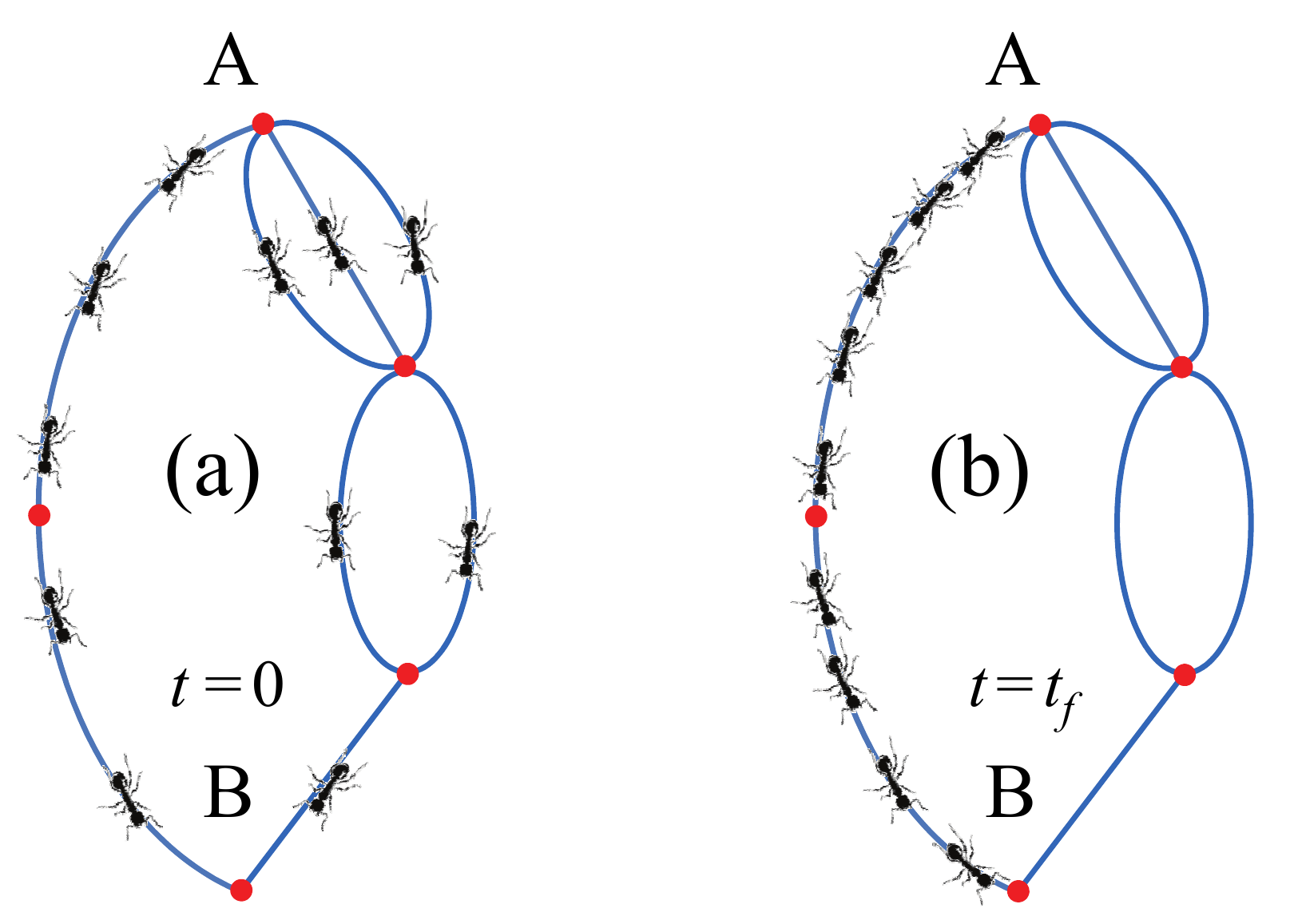}
\includegraphics[angle=0,width=7cm]{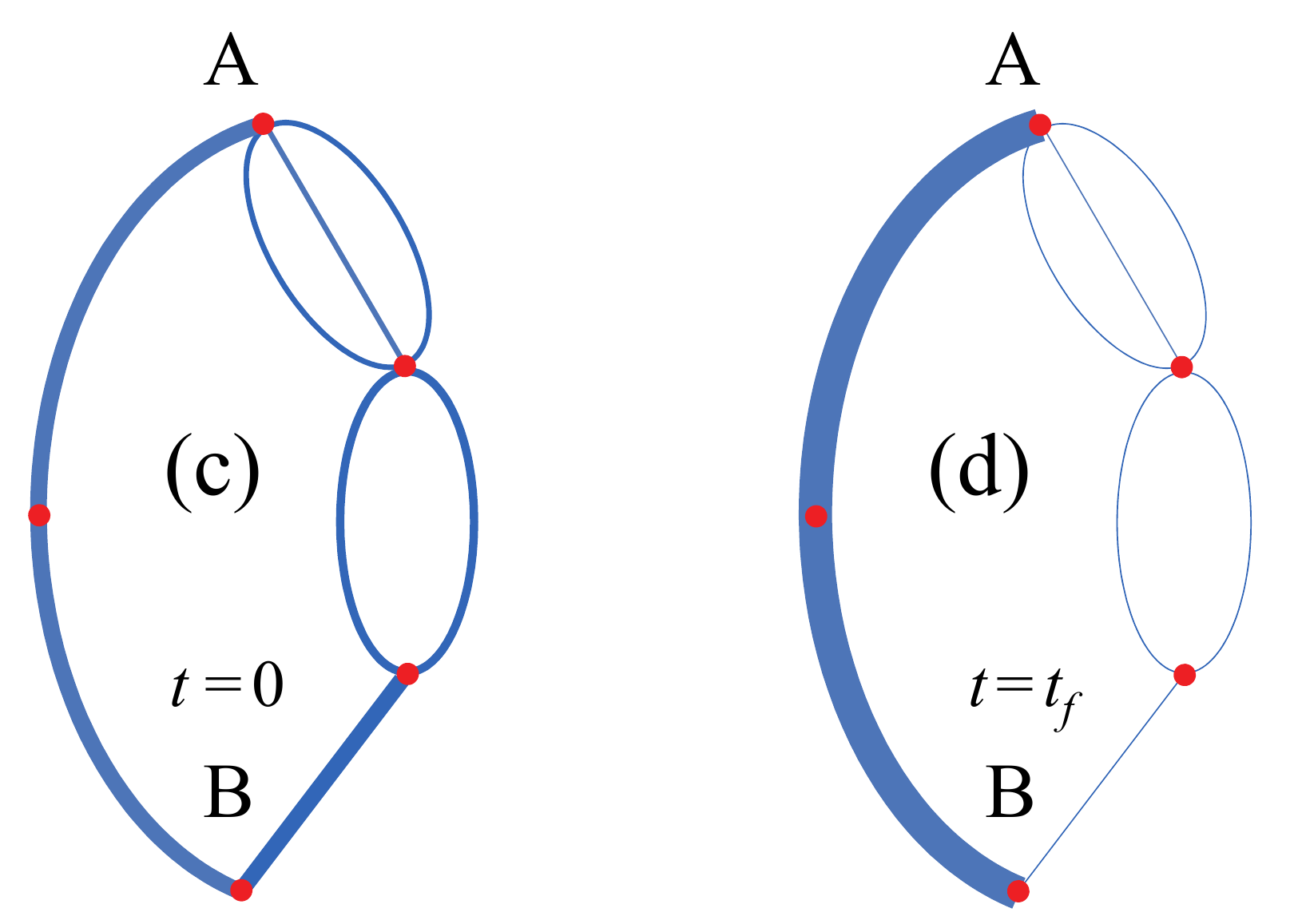}
\caption{(Color online) Schematics of distributions of moving agents (ants) (a), (b) and currents (c), (d) at the initial $t=0$ and final $t=t_f$ moments of time. These distributions correspond to the results reported in Fig. \ref{fig4}. The current strength in (c), (d) is represented by the edge thickness. }
\label{fig5}
\end{center}
\end{figure}

Fig. \ref{fig4} shows the results of both approaches. As can be seen, the correct solution is found by both methods. Moreover, at longer times, the order of the curves in Fig. \ref{fig3}(a) and (b) is the same. This is again related to the similarity between the ant colony and memristive network approaches which is even more apparent from Fig. \ref{fig5}, where we plot the
real space solutions of both methods  at steady state. In Fig. \ref{fig5}(a) and (c), the initial state is shown for the
ant colony (a) and the currents (c), while in in Fig. \ref{fig5}(b) and (d) the corresponding solutions at the final time
are represented.

\section{Memcomputing with threshold-type memristive systems} \label{sec5}

Finally, we consider the more common case of memristive elements with current or voltage thresholds. Indeed, for physical reasons~\cite{diventra13b} this is the type of elements that can mainly be realized experimentally. Therefore, it is important to understand how the threshold-type switching modifies the results we have described above. For this purpose, we perform simulations similar to those in Sec. \ref{sec6} by considering, however, a current-controlled memristive system with threshold described by Eqs. (\ref{Icontr5}), (\ref{Icontr6}).

\begin{figure}[tb]
\begin{center}
\includegraphics[angle=0,width=7cm]{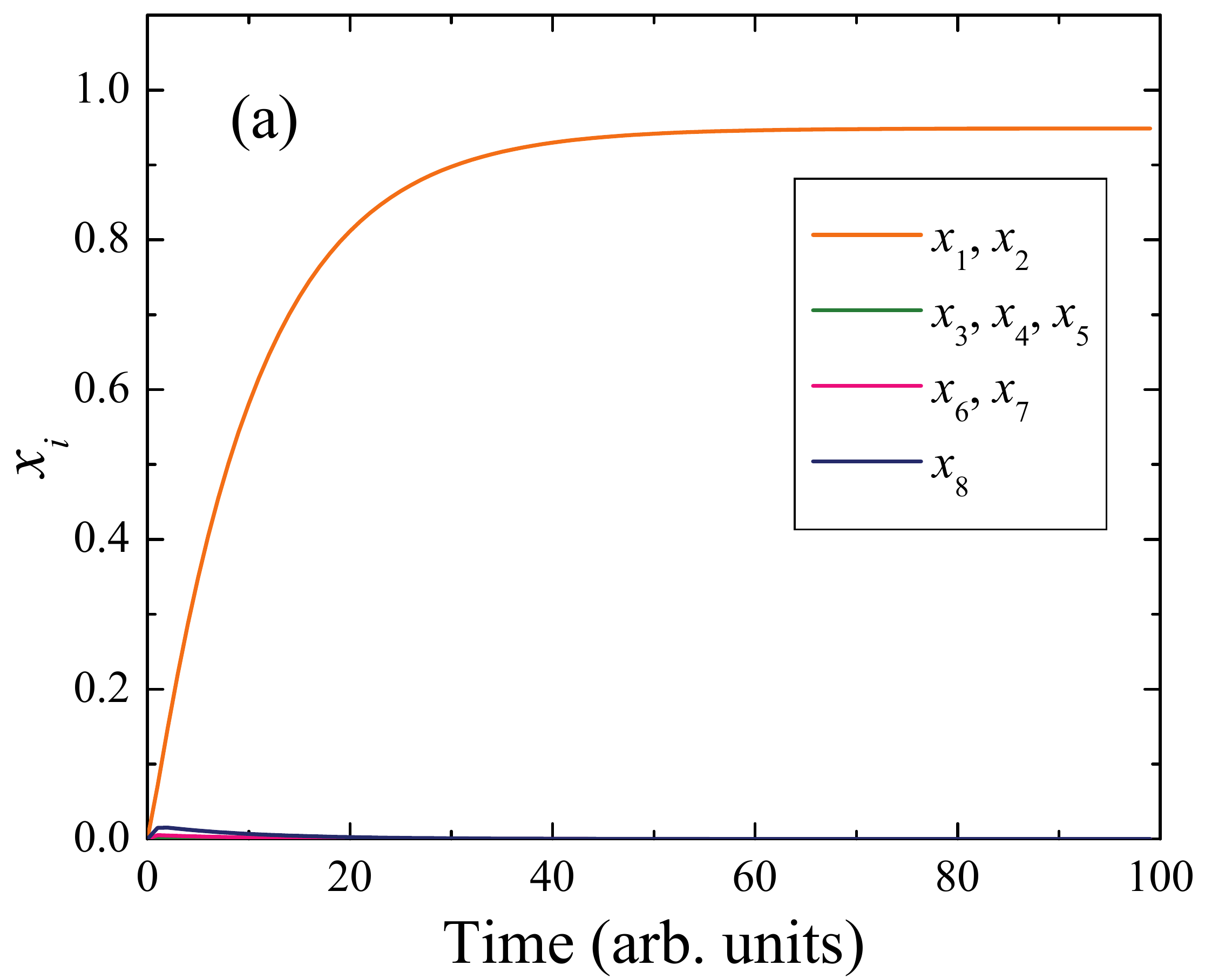}
\includegraphics[angle=0,width=7cm]{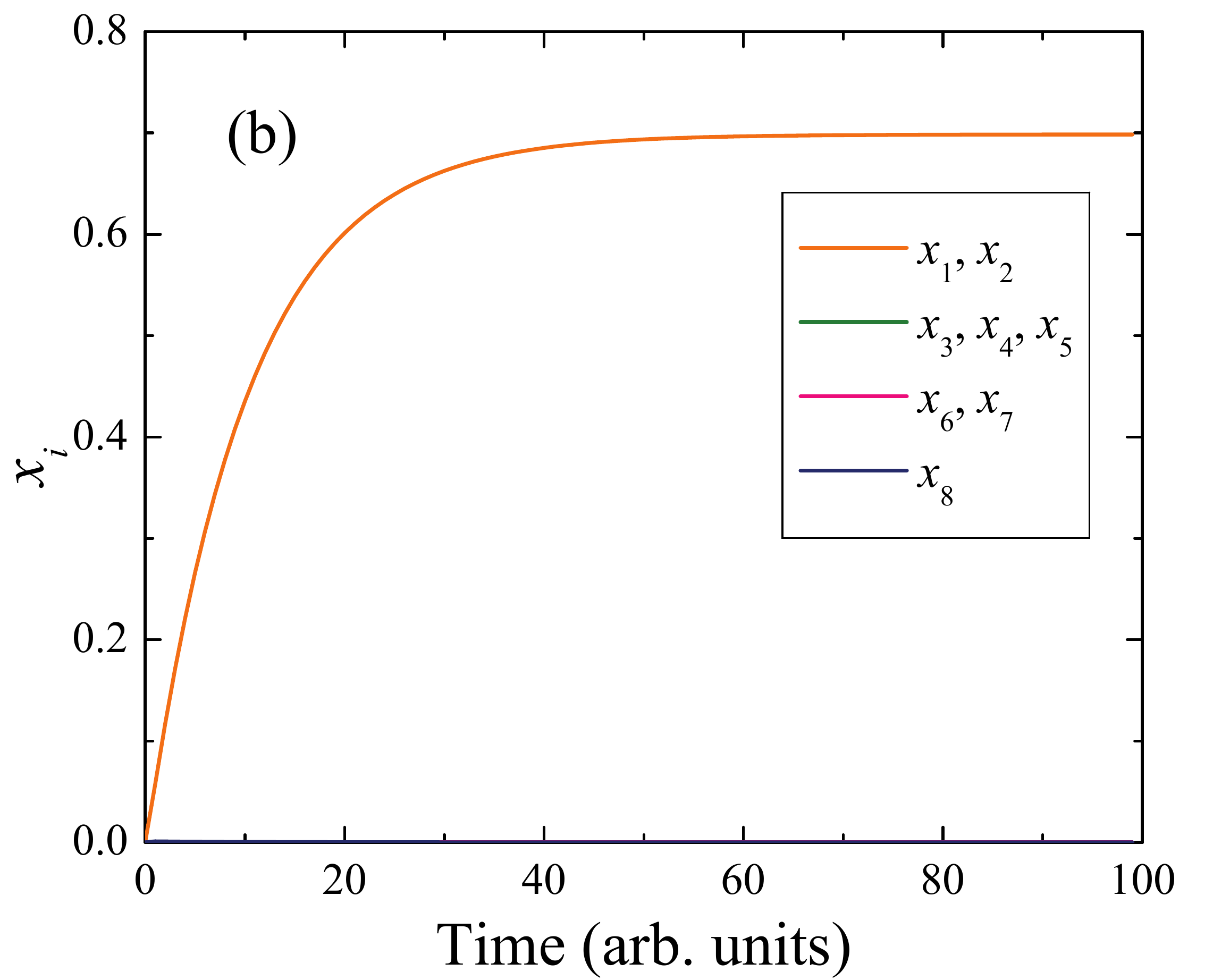}
\caption{(Color online) Dynamics of the internal state variables of memristive systems in the network shown in Fig. \ref{fig3}(b) subjected to a constant current. $I_t=0.005$ A in (a) and $0.03$ A in (b). All other parameters are
the same as in Fig. \ref{fig4}(b).}
\label{fig6}
\end{center}
\end{figure}

Results of a couple of selected simulations (representing smaller and larger threshold voltages) are shown in Fig. \ref{fig6}. In particular, Fig. \ref{fig6} shows that the correct solution is found by threshold-type memristive systems as well. Moreover, it is clear from these figures that the more realistic threshold-type memristive systems
are actually better for this type of optimization problems since the current threshold cuts off the dynamics of those memristive systems in the network that are subject to weaker currents. Additionally, one can notice a reduction of the steady-state value of $x_1$ with increase of the threshold current. This is a direct consequence of the model given by Eqs. (\ref{Icontr5}), (\ref{Icontr6}): the effective switching rate at a given current decreases with increasing $I_t$. If needed, in hardware implementations this could be easily compensated by a stronger applied current.

\section{Conclusions} \label{sec6}

In conclusion, we have shown the connection between swarm intelligence algorithms and memcomputing, namely computing with and in memory.
By focusing on the archetypal ant colony algorithm and the memristive network realization of
memcomputing, we have shown both analytically and numerically that the two approaches find the same
shortest path solutions.

We emphasize that ant colony optimization and memcomputing approaches share a lot of similarities but they are not exactly the same.
While the artificial ant dynamics is dictated by local edge selection rules, the electron current flow in memristive network
is determined by the solution of Kirchoff's laws for the entire circuit. At the same time, while the pheromone level update
depends on the total specific path length, the change of the memristive system state depends on the local current. Therefore,
as a consequence of a weaker sensitivity of memcomputing to the specific path length, one can expect a higher probability of
approximate solutions in the memcomputing approach. In fact, like in the case of ant colony approaches \cite{dorigo1997ant}, not all
correct solutions may be found with all possible memristive networks. The percentage of such solutions will depend on the topology of the network and parameters of
the memristive elements. However, in all cases considered in this paper both approaches have provided the same correct solutions, and we leave the study of possible deviation from ideal perfomance for future work.

Importantly, unlike the ant colony approach, which requires multiple stochastic steps,
memcomputing, due to its massive, intrinsic parallelism has the capability of finding the solution in one deterministic step.
This, combined with the fact that memristive systems, as those we discussed in this paper, can be easily realized
in the lab, paves the way for hardware implementations of several optimization problems that are currently solved
using swarm intelligence approaches. These implementations can then have a large impact in several practical
machine learning
problems related to urban planning, scheduling, robotics, and so on.

\section*{Acknowledgements}

This work has been partially supported by NSF grant ECCS-1202383 and the Center for
Magnetic Recording Research at UCSD. The hospitality of the Aspen Center for Physics (supported by NSF grant PHY-1066293), where part of this work was done, is also acknowledged.

\bibliographystyle{IEEEtran}
\bibliography{memcapacitor}
\end{document}